%% file: persevere.tex
\documentclass[a4paper,final]{article}
\usepackage{arxiv}
\usepackage{doi}
\usepackage{xcolor}
\definecolor{rseblue}{RGB}{2,72,142}
\hypersetup{colorlinks=true,allcolors=rseblue}
\usepackage{enumitem}
\usepackage{draftwatermark}
\SetWatermarkScale{0.45}
\SetWatermarkAngle{90}
\SetWatermarkHorCenter{1cm}
\SetWatermarkColor{red!25}
\usepackage{natbib}
\usepackage{booktabs}
\usepackage{aofigures}

\begin{document}
\title{Artificial Organisations}

\author{
  William Waites \\
  University of Southampton\\
  \texttt{\href{mailto:w.waites@soton.ac.uk}{w.waites@soton.ac.uk}}\\\texttt{\href{mailto:ww@groovy.net}{ww@groovy.net}}
}
\rhead{\scshape Artificial Organisations - \today}

\maketitle

\begin{abstract}
Alignment research focuses on making individual AI systems reliable. Human institutions achieve reliable collective behaviour differently—they mitigate the risk posed by misaligned individuals through organisational structure. Multi-agent AI systems should follow this institutional model: using compartmentalisation and adversarial review to achieve reliable outcomes through architectural design rather than assuming individual alignment.

We demonstrate this approach through the Perseverance Composition Engine, a multi-agent system for document composition. The Composer drafts text, the Corroborator verifies factual substantiation with full source access, and the Critic evaluates argumentative quality without access to sources—information asymmetry enforced by system architecture. This creates layered verification: the Corroborator detects unsupported claims, whilst the Critic independently assesses coherence and completeness. Observations from 474 composition tasks—discrete cycles of drafting, verification, and evaluation—exhibit patterns consistent with the institutional hypothesis. The verification agent detected fabrication in 52\% of submitted drafts. Iterative feedback between compartmentalised roles produced 79\% quality improvement over 4.3 iterations on average. When assigned impossible tasks requiring fabricated content, this iteration enabled progression from attempted fabrication toward honest refusal with alternative proposals—behaviour neither instructed nor individually incentivised. These findings motivate controlled investigation of whether architectural enforcement produces reliable outcomes from unreliable components.

This positions organisational theory as a productive framework for multi-agent AI safety. By implementing verification and evaluation as structural properties enforced through information compartmentalisation, institutional design offers a route to reliable collective behaviour from unreliable individual components.
\end{abstract}

\keywords{artificial organisations, multi-agent systems, organisational theory, institutional design, transactive memory, information compartmentalisation, epistemic integrity, model organisms, distributed cognition, institutional memory}

\clearpage
\input{0_acknowledgements}
\clearpage
\input{1_introduction}
\input{2_background}

\input{3_system}

\input{4_verification}
\input{5_honest_refusal}
\input{6_results}
\input{7_limitations}
\input{8_conclusion}

\subsection{Data and Source Code Availability}
Source code is available at \url{https://codeberg.org/wwaites/persevere}. All notes, drafts, composition project artefacts and so forth underlying this document are available at \url{https://codeberg.org/wwaites/persevere-data}.

\setlength{\bibsep}{0pt}
\bibliographystyle{abbrvnat}
\bibliography{multi-agent}

\clearpage
\input{9_appendix}

\end{document}

%% file: 0_acknowledgements.tex
\section*{Author Contributions and Acknowledgements}

This section appears at the beginning of the manuscript rather than at its customary end. The reason is straightforward: the Perseverance Composition Engine (PCE) composed this paper about itself, and an honest account of that process must precede the reading rather than follow it.

\subsection*{A Note on the Editorial Process}

PCE drafted this document from its own source code and informal notes provided by the authors. However, the system did not produce this paper autonomously. The composition required sustained human direction across approximately 500 distinct composition projects, involving strategic guidance, error correction, and---critically---the identification of which insights warranted attention and how they should be framed.

This editorial relationship presents an unusual challenge for scholarly attribution. The system synthesised and articulated material that no single human author wrote sentence by sentence, yet the conceptual direction, theoretical framing, and evaluative judgements remained human responsibilities throughout. Recent work on semi-autonomous mathematics discovery reaches a similar conclusion: that human authorship remains necessary because it entails accountability not only for technical correctness but also for expositional integrity and the correctness of attributions---responsibilities that only humans can bear \citep{feng2026erdos}. We place these acknowledgements first so that readers may calibrate their interpretation of what follows: a document composed by the system it describes, under the guidance and oversight of its human authors.

\subsection*{Contributions}

WW designed and implemented PCE, provided strategic direction, and developed the theoretical framework. As senior author, WW directed the composition process: instructing the system, checking the output, correcting errors, and resolving disputes between agents, soliciting of external feedback, as well as very light editing of the final draft. This oversight extended beyond factual grounding to encompass the identification of significant insights, distinguishing them from incidental details, and the overall framing of the work.

PCE composed the text from the source materials provided by the humans.

\subsection*{Acknowledgements}

PJ contributed insights from the literature and theoretical framing, conducted software testing and provided feedback, and performed the literature search that led to the case study presented in \S\ref{sec:refusal}.

We thank ABC, CT, DS, MJB, MPF, PS and SM for valuable discussion.

%% file: 1_introduction.tex
\section{Introduction}

The alignment research programme has focused predominantly on individual AI systems: training single agents to be helpful, harmless, and honest \citep{bai2022constitutional}, ensuring reward models capture human intentions \citep{christiano2017deep, leike2018scalable}, detecting and preventing deceptive alignment in individual models \citep{hubinger2019risks, hubinger2024sleeper}. This focus is natural—single agents are the unit of deployment and the unit of training. But it represents an individualist assumption that human institutions long ago abandoned. Human organisations do not assume that individuals are reliably aligned with collective goals. They assume the opposite: that individuals have bounded rationality \citep{march1958organizations}, conflicting interests, and limited oversight capacity. Institutions achieve reliable collective behaviour through structure—separation of duties, adversarial review, information compartmentalisation, audit cycles—not through perfecting individual character. Organisational theory treats this structure as an information-processing response to bounded rationality \citep{march1958organizations, galbraith1974designing}: separation of duties distributes decision-making, compartmentalisation isolates risks, and audit cycles enable detection and correction. The question this paper poses is whether multi-agent AI systems can follow the same design logic. We demonstrate both a theoretical reframing and an empirical test: multi-agent AI systems can achieve reliable collective behaviour from individually unreliable components through institutional structure rather than assuming alignment through training.

This reframing translates organisational vocabulary into architectural constraints. Where human institutions establish separation of duties through policies and procedures, multi-agent systems can enforce it through information access controls. Where organisations rely on adversarial review—audit functions, peer review, adversarial legal proceedings—prior work has proposed debate mechanisms where adversarial agents improve reasoning and factuality \citep{irving2018ai, du2023debate}. This paper extends that logic by implementing information compartmentalisation as an architectural constraint: verification and evaluation roles operate with structurally distinct information access, enforced by the system rather than instructed to agents. Where human groups develop organisational memory systems that retain and retrieve information across members \citep{walsh1991organizational,stein1995actualizing}, multi-agent architectures can implement role-specific memory with differentiated access rights. The Perseverance Composition Engine demonstrates this translation in document composition through layered verification: a Composer drafts text, a Corroborator verifies factual substantiation with full source access, and a Critic evaluates argumentative quality without access to sources. Information asymmetry is enforced architecturally, not instructed. The Corroborator detects unsupported claims; the Critic independently assesses coherence and completeness. This creates verification that no individual agent could guarantee—the institutional structure produces the checking function.

Across 474 completed projects---discrete, small composition or curation tasks---this institutional architecture produced measurable behaviours relevant to AI safety. The system achieved a 69\% project completion rate, with successful projects requiring a mean of 4.3 iterations between drafting, verification, and evaluation to reach convergence. Quality scores improved by 78.85\% on average from initial submission to final acceptance, at a mean cost of \$0.29 per completed project—demonstrating economic feasibility of structured verification at scale. Across documents reviewed for substantiation, the verification agent classified 52\% as fabricated, requiring iterative revision toward full substantiation. Two observations illustrate the mechanisms at work. During self-documentation of the system's own capabilities, verification correctly rejected six consecutive drafts, each time distinguishing implemented architectural features from plausible but unsubstantiated claims. When assigned an impossible task requiring literature synthesis from raw search transcripts, the system progressed over five iterations from attempted fabrication toward refusal with alternative proposals—behaviour neither explicitly instructed nor individually incentivised. The progression occurred through iterative feedback between verification and evaluation roles operating under information compartmentalisation. We position artificial organisations as model systems \citep{hubinger2024sleeper} for studying how institutional structure shapes collective behaviour in AI systems, extending the model organisms methodology from individual to collective safety properties. This operational corpus includes the composition of this paper itself, creating methodological reflexivity we address in \S\ref{sec:limitations}. Detailed aggregate statistics appear in \S\ref{sec:results}.

This paper documents the design, implementation, and operational behaviour of this institutional architecture. \S\ref{sec:background} reviews the organisational theory foundations and positions the work relative to contemporary multi-agent systems research. \S\ref{sec:architecture} describes the seven-agent system architecture, showing how separation of duties, information compartmentalisation, and adversarial review translate into computational constraints enforced by system architecture rather than agent instruction. \S\ref{sec:verification} and \S\ref{sec:refusal} present the two case studies: verification rigour under self-documentation and progression toward refusal under impossible task constraints. \S\ref{sec:results} provides aggregate statistics across 474 projects, documenting fabrication classification rates, quality improvement trajectories, iteration counts, and operational costs. \S\ref{sec:limitations} acknowledges constraints: the case studies are illustrative observations from a working system, not controlled experiments; the system operates in a bounded domain with reflexive elements during self-documentation; and all results were produced with a single underlying model family (Claude 4.5). \S\ref{sec:conclusion} discusses implications for multi-agent AI safety: how institutional design offers a complementary approach to individual alignment, treating unreliable components as the baseline condition and achieving reliability through structure rather than assuming it through training.

%% file: 2_background.tex

\section{Background}
\label{sec:background}

\input{2.1_rationality}
\input{2.2_adversarial}
\input{2.3_model_systems}
\input{2.4_related}

%% file: 2.1_rationality.tex
\subsection{Bounded Rationality and Institutional Structure}
\label{subsec:bounded-rationality}

Organisations exist because individuals cannot solve complex problems alone. \citet{march1958organizations} established that human decision-makers operate under bounded rationality: constrained attention, limited memory, and finite computational capacity make it impossible for any individual to process all relevant information, evaluate all alternatives, or predict all consequences. Organisations respond by decomposing complex problems into tractable subproblems, distributing information to appropriate decision-makers, and coordinating distributed activities through structure. The question is not whether individuals are rational, but how structure compensates for the fact that they are not.

\citet{simon1962architecture} demonstrated that hierarchical structure is not administrative convention but an evolutionary solution to complexity. His watchmaker parable contrasts two approaches to assembling a watch with one thousand parts. The first craftsman works with flat structure, assembling all parts in a single sequence; any interruption forces recommencement from the beginning. The second uses hierarchical assembly, creating stable intermediate subassemblies of ten parts each, then assembling ten subassemblies into the watch. Given even modest interruption probability, the hierarchical approach proves approximately four thousand times faster. The principle generalises: hierarchical decomposition with stable intermediate states dramatically reduces search spaces and enables progress despite interruption. Simon formalised this insight as \emph{nearly decomposable systems}—hierarchies where within-subsystem interactions substantially exceed between-subsystem interactions. Such systems exhibit approximate independence in the short run whilst long-run behaviour depends only on aggregate inter-subsystem relationships. This explains why organisations function despite imperfect communication and why modular analysis remains tractable.

This perspective reframes organisational structure as epistemic architecture. How labour is divided determines what information each role encounters; reporting relationships govern information flow; standard operating procedures crystallise organisational knowledge into executable routines. Structure shapes not only what an organisation does but what it can perceive, retain, and decide. \citet{galbraith1974designing} elaborated this framework by characterising organisations as information-processing systems responding to environmental uncertainty. Routine tasks permit hierarchical structures with constrained information channels. Tasks marked by uncertainty and interdependence demand greater capacity through lateral relations, slack resources, or enhanced information systems. Galbraith's insight is that optimal structure is contingent: it must match the information-processing demands imposed by the task environment.

Individual LLM agents operating under bounded context windows face cognitive constraints directly analogous to human bounded rationality, making organisational structure—not individual training—the appropriate design lever for collective reliability. The architecture governing task routing, context sharing, and output aggregation determines what each agent perceives and how information flows between roles. Design decisions about specialisation, information compartmentalisation, and coordination mechanisms fundamentally shape collective capabilities \citep{wu2023autogen, du2023debate}.

%% file: 2.2_adversarial.tex
\subsection{Adversarial Structure and Verification}
\label{sec:adversarial}

Human institutions do not assume that individuals are perfectly reliable. Instead, they design structural safeguards: auditors verify financial statements prepared by others, peer reviewers evaluate manuscripts without knowing authors' identities, and intelligence agencies compartmentalise classified information to limit damage from compromised individuals. The principle is consistent across domains—separation of roles creates checking functions that no single actor can circumvent. Multi-agent AI systems can implement this principle more rigorously than human organisations, transforming policy-based restrictions into architectural constraints.

\subsubsection{Debate as Verification Framework}

\citet{irving2018ai} established debate as a mechanism for scalable oversight, proving that debate between competing agents can answer any question in PSPACE and demonstrating empirically that adversarial selection improves verification tractability. In their MNIST classification experiments, a sparse classifier improved from 59.4\% to 88.9\% accuracy when shown only six pixels selected adversarially by debating agents, and from 48.2\% to 85.2\% with just four pixels. The debaters identify which evidence matters for correct classification; the judge need not examine the full input space.

\citet{du2023debate} demonstrated that multiagent debate improves factual accuracy in contemporary language models. In their experiments, problems where all models initially produced incorrect answers frequently converged to correct solutions through iterative debate, with accuracy improvements ranging from 8\% to over 30\% depending on task complexity and model capability. These results establish that adversarial interaction between language models can improve output quality beyond what individual models achieve in isolation.

\citet{harrasse2024d3} introduced Debate, Deliberate, Decide (D3), a structured multi-agent evaluation framework using role-specialized Advocates, Judge, and Jurors to reduce bias in LLM evaluation. D3 offers two protocols: Multi-Advocate One-Round (MORE) achieving 85.1\% accuracy on MT-Bench, and Single-Advocate Multi-Round (SAMRE) reaching 86.3\% whilst reducing token costs 40\% through budgeted stopping. Across three benchmarks, D3 outperforms existing multi-agent approaches through structured debate and explicit cost-accuracy trade-offs.

\subsubsection{Information Compartmentalisation}

Adversarial verification—the use of competing or complementary agents in verification roles—depends on information compartmentalisation to position each agent for independent assessment. Because we want different forms of verification, we restrict information access based on verification role; the restriction enables the verification function. Both debate and compartmentalisation operate on the same principle: verification improves when agents have constrained information access that focuses them on specific aspects of the problem.

Human organisations recognise compartmentalisation's value but must enforce it through policy and cultural norms. Financial institutions erect Chinese walls between advisory and trading divisions \citep{herzel1978chinese}, requiring employees not to share information across the barrier. Peer review systems request author anonymisation and, in double-blind review, instruct reviewers not to identify themselves, attempting to prevent prestige bias \citep{tomkins2017reviewer}. Clinical trials employ double-blind protocols—blinding both investigators and participants—to prevent systematic bias from either party's knowledge of treatment assignment \citep{schulz2002blinding}. In each case, compartmentalisation depends on individual compliance—reviewers could violate anonymity, authors could fail to anonymise adequately, and employees could share information despite prohibitions.

Artificial multi-agent systems enable a qualitatively different approach: architectural enforcement through code-level access restrictions. Where human organisations must rely on training, monitoring, and sanctions to maintain compartmentalisation, an AI agent without programmatic access to certain documents cannot consult them regardless of instruction or incentive. Unlike human reviewers who might violate anonymity norms, an agent without programmatic access cannot, ensuring the constraint is constitutive of its capability rather than a behavioural expectation requiring oversight.

This architectural capacity addresses a fundamental tension in document verification. Effective evaluation requires both fidelity to source materials—to prevent fabrication of unsupported claims—and independence from them—to assess whether a draft synthesises sources into novel argument or merely paraphrases existing formulations. A single evaluator with unrestricted access faces conflicting demands: knowing the sources makes it difficult to judge synthesis quality independently of familiarity with source content. Instructions to maintain independence might fail under pressure or cognitive load; architectural restriction cannot be violated.

\subsubsection{Complementary Verification Roles}

Multi-agent architectures can resolve this tension through role differentiation with distinct information access. In document composition systems, one verification agent might receive both draft text and complete source materials, positioned to check whether claims are substantiated. Another might receive only the draft and assignment specifications, positioned to evaluate argumentative coherence without reference to sources. The first role addresses fabrication risk; the second addresses synthesis quality. Each agent's blind spot becomes the other's vantage point; compartmentalisation ensures neither can over-correct for the other's limitations.

This approach instantiates the institutional principle that specialised roles with differentiated information access enable verification functions that comprehensive individual evaluation cannot reliably provide. The verification problem decomposes into distinct aspects—substantiation versus synthesis, accuracy versus argumentation—with architectural constraints ensuring that each agent addresses its assigned aspect without cross-contamination from the other's perspective.

The Perseverance Composition Engine implements this structure through two verification agents: a Corroborator with access to source materials who verifies factual substantiation, and a Critic without source access who evaluates argumentative quality and communicative effectiveness. Neither agent individually assesses all dimensions of document quality; the architecture positions each to address failure modes the other cannot detect. A document must satisfy both verifications to progress—it must be simultaneously well-substantiated and well-argued.

Human organisations achieve similar functional separation through procedural rules, but the compartmentalisation remains permeable. An academic reviewer instructed not to consult external sources might do so anyway; an auditor might inappropriately communicate with those being audited. In artificial systems, information access can be architecturally definitive. The Critic agent has no mechanism to retrieve source materials; the constraint is not procedural but structural. This transforms compartmentalisation from a behavioural norm requiring enforcement into a constitutive property of the verification architecture.

This architectural approach extends to any verification task where multiple orthogonal criteria must be assessed: factual accuracy and reasoning quality, technical correctness and communicative effectiveness, compliance and innovation. Wherever such tasks arise, adversarial verification with information compartmentalisation offers a structural approach. The architecture does not require that individual agents be comprehensively capable; it positions specialised agents to address distinct aspects of a problem under information constraints that enable independent judgement of those aspects.

%% file: 2.3_model_systems.tex
\subsection{Model Systems for AI Safety}

The study of AI safety increasingly employs model organisms—systems where potential failure modes can be investigated under controlled conditions \citep{hubinger2024sleeper}. \citet{hubinger2024sleeper} demonstrated that deceptive alignment could be induced in individual language models through training interventions, creating ``sleeper agents'' that persisted despite safety training. We extend this methodology to collective behaviour: artificial organisations as model systems for studying institutional dynamics in multi-agent systems.

Where model organisms of misalignment examine individual agent failures, multi-agent systems enable investigation of organisational mechanisms. Information compartmentalisation in peer review depends on behavioural compliance—reviewers instructed not to identify authors, authors requested to anonymise submissions \citep{tomkins2017reviewer}. In contrast, artificial systems implement compartmentalisation as access control constraints: an agent without programmatic access to certain documents cannot consult them regardless of instruction. The tractability advantage parallels biological model organisms—not ecological representativeness but experimental control over causal mechanisms.

Artificial organisations differ from human institutions in degree rather than kind, but the difference matters methodologically. Communication protocols are logged rather than reconstructed from testimony. Information access reflects architectural constraints—what documents an agent can retrieve—rather than policy adherence. The Perseverance Composition Engine's separation of fact-checking (Corroborator with source access) from argumentative evaluation (Critic without sources) implements compartmentalisation as code rather than procedure. This separation applies to any verification task where multiple orthogonal criteria must be assessed—factual accuracy versus argumentative coherence, technical correctness versus communicative effectiveness. Such architectures enable investigation of how coordination failures emerge, how compartmentalisation affects fabrication rates, and how adversarial verification influences convergence trajectories.

Contemporary multi-agent frameworks like AutoGen \citep{wu2023autogen}, CAMEL \citep{li2023camel}, and MetaGPT \citep{hong2023metagpt} demonstrate that LLM-based agents can coordinate on complex tasks, whilst benchmarks like MultiAgentBench \citep{zhu2025multiagentbench} provide standardised evaluation environments. Artificial organisations as model systems complement this work by treating institutional structure as the experimental variable: does architectural enforcement of information compartmentalisation reduce fabrication? Do adversarial verification arrangements improve output quality beyond single-agent baselines? The unit of analysis shifts from individual agent capability to institutional arrangement.

Generalisability to deployment contexts remains an empirical question. Model systems reveal which institutional mechanisms function under controlled conditions; field validation establishes robustness under operational constraints. For AI safety, this methodology enables investigation of structural approaches to alignment—achieving reliable collective behaviour through institutional design rather than perfecting individual agent alignment.

%% file: 2.4_related.tex
\subsection{Related Work}

Contemporary multi-agent LLM frameworks demonstrate that foundation models can coordinate effectively on complex tasks. AutoGen provides conversable agents integrating LLMs, human inputs, and tools through unified interfaces, supporting flexible conversation patterns for task decomposition and collaborative problem-solving \citep{wu2023autogen}. CAMEL employs role-playing frameworks where AI assistant and user agents use inception prompting to generate collaborative solutions, with role-flipping mechanisms enabling perspective-taking \citep{li2023camel}. MetaGPT assigns agents specific organisational roles—product managers, architects, engineers—mirroring human software development workflows to structure code generation tasks \citep{hong2023metagpt}. These frameworks prioritise demonstrating agent capabilities through structured communication protocols and role assignment.

\citet{lamalfa2025} identify four critical gaps in current LLM-based multi-agent systems research: the social dimension of agency beyond simple role assignment, systematic principles for environment design, coordination protocols that exploit LLM capabilities whilst managing their limitations, and frameworks for measuring emergent collective behaviours. Contemporary frameworks, they argue, focus predominantly on task decomposition and capability demonstration rather than the architectural principles that ensure reliable coordination. The Perseverance Composition Engine addresses these gaps through complementary mechanisms. Social agency emerges from specialised roles with distinct epistemic access: Curator, Composer, Corroborator, Critic operate within designed information environments (document groups with visibility constraints). Coordination protocols implement adversarial-cooperative dynamics: the Corroborator verifies factual substantiation with source access whilst the Critic evaluates argumentative completeness without sources. Emergent collective behaviour manifests in iterative improvement trajectories, measured through convergence patterns in verification scores (detailed in \S\ref{sec:results}).

The distinction between policy-based and architectural enforcement matters for verification systems because it determines the locus of reliability. In policy-based systems, compartmentalisation depends on agent compliance with instructions; an agent told not to consult sources might nonetheless retrieve them if prompted differently. In PCE, information access reflects programmatic constraints: an agent without retrieval permissions for certain documents cannot consult them regardless of instruction. This architectural separation implements institutional principles from human organisations—separation of duties \citep{li2007mutually}, information compartmentalisation via ``Chinese walls'' \citep{brewer1989chinese}, and adversarial verification through debate \citep{irving2018ai}—as computational access control. Where capability-focused frameworks ask what agents can accomplish, PCE investigates how institutional structure constrains agent behaviour to achieve reliable collective outcomes from individually unreliable components. Contemporary frameworks could incorporate such architectural constraints alongside their coordination protocols; PCE demonstrates their feasibility and provides vocabulary from organisational theory for reasoning about multi-agent safety through structural design rather than capability perfection.

%% file: 3_system.tex
\section{System Architecture}
\label{sec:architecture}

\input{3.1_overview}
\input{3.2_seven_agents}
\input{3.3_information}
\input{3.4_coordination}
\input{3.5_memory}

%% file: 3.1_overview.tex
\subsection{Overview}
\label{sec:arch-overview}

The Perseverance Composition Engine operationalises the institutional design principles from \S\ref{sec:background} through a three-layer architecture that enforces information compartmentalisation as a structural property rather than a behavioural policy. Where human organisations achieve reliable collective outcomes by constraining individual behaviour through role separation and information access controls, PCE implements these constraints programmatically: agents cannot access information outside their designated scope regardless of instruction or prompt manipulation. This architectural approach reflects March and Simon's foundational insight that organisations exist because structure compensates for individual limitations—in PCE's case, the limitations of individual language models operating under epistemic confinement.

\figAOdiagram

The architecture comprises three integrated layers: a document catalogue with metadata-driven visibility constraints, a network of specialised agents with role-appropriate information access, and a workflow orchestrator managing iterative refinement (Figure~\ref{fig:agent-network}). Layer 1, the document catalogue, organises documents into groups and records metadata. Each document carries a visibility field that determines which agents may access it. Six visibility levels regulate information flow: PUBLIC documents are universally accessible; CANDIDATE documents contain source materials visible to Composer, Corroborator, and Curator but hidden from Critic; DRAFT documents contain composition outputs; FEEDBACK documents contain evaluation reports from Corroborator and Critic; CRITIC documents are visible only to the reviewing agent; ARCHIVE documents represent historical records. Access control operates through specialised document-listing functions (\texttt{public\_document\_list}, \texttt{candidate\_document\_list}, \texttt{critic\_document\_list}) that agents invoke to retrieve documents. Each agent is provisioned only with functions matching its institutional role, making information access a constitutive architectural property. An agent without the retrieval function for CANDIDATE documents cannot consult them through any means short of system modification.

Layer 2, the agent network, comprises six core agents—Concierge, Commutator, Curator, Composer, Corroborator, and Critic—each provisioned with distinct information access profiles and functional capabilities appropriate to their verification roles within the institutional structure. An auxiliary Compressor agent operates outside the main workflow to manage context window constraints. The key architectural separation occurs between Corroborator and Critic: both evaluate draft quality, but Corroborator accesses source materials to verify factual substantiation whilst Critic operates without source visibility to provide independent assessment of argumentative completeness. This differential access implements the institutional principle of separation of duties, ensuring that drafts satisfy independent filters operating under complementary epistemic constraints. Detailed agent descriptions and their institutional roles appear in \S\ref{sec:seven-agents}

Layer 3, the workflow orchestrator, manages execution through a directed state machine (PerseveranceGraph) that sequences agent invocations: Commutator triages incoming projects---discrete pieces of work such as drafting a short document or section of a document, optional Curator enrichment occurs, Composer generates drafts, Corroborator verifies substantiation, Critic evaluates quality, and if scores fall below the convergence threshold, feedback returns to Composer for revision. The GraphState data structure propagated through workflow nodes maintains iteration count and feedback accumulation, enabling Composer to build on prior revisions whilst Critic evaluates each draft independently. The workflow operates asynchronously, ensuring that long-running composition projects never block user interaction through the Concierge. Convergence occurs when Critic assigns scores exceeding the acceptance threshold following Corroborator verification that all claims are substantiated by source materials.

This architecture differs fundamentally from policy-based systems where access control depends on agent compliance with instructions. Information compartmentalisation becomes architectural enforcement: the Critic cannot verify claims against sources because the function to retrieve CANDIDATE documents does not exist in its provisioned tool suite. The workflow ensures layered verification—substantiation precedes quality assessment—with each layer operating under designed informational constraints. This decomposition addresses epistemic confinement by distributing review across agents with complementary capabilities and limitations, implementing institutional principles of reliable collective behaviour through structural design rather than individual agent perfection.

%% file: 3.2_seven_agents.tex
\subsection{The Seven Agents}
\label{sec:seven-agents}

Building on the three-layer architecture described in \S\ref{sec:arch-overview}, the agent network comprises seven specialised roles with distinct information access profiles that instantiate institutional design principles through architectural enforcement. Each agent operates with fixed capabilities and retrieval permissions appropriate to its verification function within the collaborative workflow.

\textbf{Concierge} manages the boundary between user and system, transforming underspecified requests into formal task definitions before workflow execution begins. It has unrestricted document access and is the only agent permitted to request user clarifications, instantiating the principle of requirements elicitation under bounded rationality: downstream agents cannot infer unstated user knowledge and require explicit specifications to optimise within their specialised domains. The Concierge's boundary role separates client-facing coordination from internal production work, ensuring that composition agents receive well-defined problems rather than ambiguous mandates that would require iterative user consultation during execution.

\textbf{Commutator} receives specifications from the Concierge and determines optimal routing through the composition pipeline based on semantic analysis of project goals and available materials. It accesses the document database but maintains no conversational state between projects, performing fresh triage on each remit. The Commutator routes work to the Curator for metadata enrichment, to the Composer for drafting, or to the Corroborator for verification without new composition. This role instantiates centralised coordination under separation of concerns: different project types require different pipeline configurations, and unified routing prevents agents from making suboptimal local decisions about workflow sequencing. The Commutator also manages feedback loops when revision is required, determining whether to return work to the Composer or escalate through task decomposition.

\textbf{Curator} maintains institutional memory through the persistent document database, organising materials with comprehensive metadata including source type, visibility constraints, document classification, authorship, and timestamps. The Curator accesses and modifies the database whilst remaining stateless with respect to GraphState between projects, enabling fresh analysis without conversational history retention. This instantiates organisational memory as externally stored, searchable knowledge structures rather than accumulated unstructured records. When invoked by the Commutator, the Curator provides document summarisation, classification of new materials, cross-referencing with prior work, and identification of knowledge gaps. These analyses transform isolated projects into situated instances of accumulated learning, enabling iterative capability development across the organisation's operational history.

\textbf{Composer} generates draft documents from CANDIDATE visibility source materials and task specifications, optimising for synthesis rather than verification. The Composer's message history in GraphState accumulates feedback from both Corroborator and Critic, enabling iterative refinement toward dual quality criteria without requiring the Composer itself to simultaneously generate and verify content. This instantiates specialised information processing: synthesis and verification impose conflicting cognitive demands that cannot be jointly optimised within a single agent. The Composer represents creative production capacity that institutional structure must both enable and constrain—it produces fluent multi-source synthesis whilst downstream agents enforce independent quality gates on factual accuracy and argumentative completeness.

\textbf{Corroborator} performs systematic factual verification with unrestricted access to both draft documents and CANDIDATE source materials, enabling direct comparison between claims and evidence. The Corroborator identifies fabrications, unsupported inferences, and evidential gaps, providing detailed feedback specifying which claims lack substantiation. This instantiates adversarial verification through information advantage: the Corroborator evaluates claims against an independent evidential standard without stake in the draft's success, addressing the confirmation bias problem where generators interpret their own output generously. The Corroborator gates on factual accuracy, determining whether drafts meet substantiation standards before quality review proceeds. When verification fails, the draft returns to the Composer with specific remediation guidance rather than advancing to the Critic.

\textbf{Critic} evaluates drafts against task specifications under architecturally enforced information compartmentalisation: the Critic cannot access source materials. Receiving only draft text and specifications, the Critic assesses whether the draft addresses requirements, maintains internal coherence, presents claims clearly, and meets venue standards. This instantiates blind review principles where source access creates availability bias—evaluators reason backward from evidence existence to claim plausibility rather than evaluating communicative effectiveness for uninformed readers. The Critic's retrieval functions exclude CANDIDATE documents regardless of instruction, making compartmentalisation constitutive rather than advisory. The Critic gates on argumentative completeness and venue appropriateness under complementary epistemic constraints to the Corroborator, ensuring drafts satisfy independent quality filters before external submission.

\textbf{Compressor} manages semantic compression of GraphState message histories when token consumption approaches model context limits, producing condensed representations that preserve essential feedback whilst discarding verbatim exchanges. An orchestrator monitors token counts, triggering compression automatically at 75\% capacity. The Compressor operates statelessly on provided histories, replacing verbose accumulation with structured summaries of claims, evidence, and revision patterns. This instantiates efficient information retention under resource constraints: principled summarisation enables sustained operation where full-fidelity storage would cause workflow termination. Compression also forces explicit claim representation—assertions requiring extensive elaboration often indicate unclear thinking or inadequate grounding. Token tracking signals the Commutator when budget constraints approach, enabling work prioritisation or task decomposition before resource exhaustion.

This specialisation with differential information access implements the institutional principle that reliable collective behaviour emerges from structural constraints rather than individual perfection. The Composer need not verify every claim because the Corroborator provides factual oversight with source access. The Corroborator need not evaluate argumentative quality because the Critic independently assesses communicative effectiveness. The Critic need not verify sources because the Corroborator gates on substantiation before quality review. Each agent optimises within designed informational constraints, and their coordinated operation produces outputs satisfying multiple independent quality criteria through layered verification rather than individual omniscience.

%% file: 3.3_information.tex
\subsection{Information Compartmentalisation}
\label{sec:information}

The institutional design principle that reliable collective behaviour emerges from structural constraints rather than individual perfection \citep{march1958organizations} finds concrete expression in the Perseverance Composition Engine through architectural enforcement of information compartmentalisation. Where the preceding section described seven agents with distinct roles, this section examines how differential information access between the Corroborator and Critic enables layered verification that addresses incompatible epistemic requirements through designed separation rather than individual discipline.

Information compartmentalisation in human organisations typically relies on policy-based restrictions that remain vulnerable to breach. In corporate settings, Chinese walls between divisions depend on employee compliance \citep{herzel1978chinese}; in peer review, blinding procedures can be circumvented through author identification \citep{tomkins2017reviewer}. Such policies establish norms but cannot guarantee epistemic separation when individuals possess means to access restricted information. PCE implements compartmentalisation differently: through code-level access restrictions that determine which document-listing functions each agent receives during instantiation. The Critic agent is provisioned with \texttt{public\_document\_list}, \texttt{draft\_document\_list}, and \texttt{feedback\_document\_list}, but not \texttt{candidate\_document\_list}---the retrieval function required to access source materials. This is not a guideline; it is a structural limitation encoded in the agent's tool configuration. The Critic cannot verify claims against sources because the architectural layer makes source documents unretrievable regardless of instructions or reasoning. Compartmentalisation becomes constitutive rather than advisory.

This architectural enforcement enables layered verification through agents with complementary epistemic constraints. The Corroborator accesses both draft documents and CANDIDATE visibility sources, enabling direct comparison between claims and evidence. Its verification detects fabrication, unsupported inferences, and evidential gaps by systematically checking whether claims find substantiation in available materials \citep{du2023debate}. The Corroborator gates on factual accuracy before drafts advance to quality review.

The Critic, receiving only draft text and task specifications, evaluates whether the draft addresses requirements, maintains internal coherence, and presents arguments clearly. This epistemic constraint addresses a distinct failure mode: when evaluators know that supporting evidence exists, they may assess claims as adequately supported even when the draft itself fails to establish connections persuasively. The Critic evaluates drafts without knowing whether supporting evidence exists, forcing assessment of argumentative completeness in the text itself rather than confidence in underlying evidence. This separation between verification with source access and verification without source access prevents unwarranted deference to claims based on evidence availability rather than textual exposition quality.

Layered verification addresses distinct failure modes through compositional assessment: the coordinated application of multiple verification functions, each operating under designed informational constraints that prevent conflation of incompatible epistemic requirements. The Corroborator's evidential access enables detection of fabrication---claims lacking source support---which requires comparing draft assertions against materials. The Critic's evidential isolation enables detection of argumentative incompleteness---claims inadequately developed for uninformed readers---which requires evaluating the draft as self-contained exposition. These verification functions cannot be jointly optimised within a single agent: source access creates confirmation bias where evaluators reason from evidence existence to claim adequacy, whilst source isolation prevents substantiation checking. By distributing these functions across agents whose architectural constraints enforce rather than recommend epistemic separation, the system achieves comprehensive verification through compositional operation: multiple partial perspectives combine to approximate evaluation that no single perspective could provide whilst maintaining both evidential access and evaluative distance simultaneously.

This architectural compartmentalisation operationalises the institutional model by distributing incompatible requirements across agents whose information access prevents conflation of distinct verification functions \citep{march1958organizations,du2023debate}. Structure compensates for individual limitations, enabling reliable collective verification without requiring individual agents to maintain epistemic discipline against natural tendencies toward confirmation or to achieve simultaneous optimisation of incompatible goals.

%% file: 3.4_coordination.tex
\subsection{Coordination and Iteration}
\label{sec:coordination}

Following the architectural separation described in \S\ref{sec:information}, where the Corroborator and Critic operate under complementary epistemic constraints, coordination between agents proceeds through an iterative feedback loop that enables progressive refinement. The system maintains workflow state through GraphState, a persistent data structure recording draft documents, verification verdicts, quality assessments, iteration count, and accumulated feedback. This state preservation enables the three verification functions---composition, substantiation checking, and quality assessment---to operate sequentially whilst maintaining coherent progress toward convergence.

The workflow implements a three-stage cycle. The Composer generates draft text from source materials and task specifications. The Corroborator verifies claims against available sources, issuing either SUBSTANTIATED (enabling progression) or FABRICATED (returning detailed feedback on evidential gaps and unsupported assertions). Substantiated drafts advance to the Critic, which evaluates argumentative quality, coherence, and specification compliance, assigning a score from 0 to 100 with qualitative guidance. Drafts scoring below the threshold return to the Composer with accumulated feedback from both verification stages. Each iteration preserves the complete feedback history: the Composer receives guidance on both evidential gaps and argumentative weaknesses, enabling simultaneous correction of fabrication and exposition deficiencies.

The process continues until the Critic's score reaches $\tau = 85$, the operationally specified convergence threshold. This value represents a pragmatic operating point: sufficiently demanding to reject drafts with substantial deficiencies, yet achievable within reasonable iteration budgets. The convergence criterion implements institutional iteration through structured feedback and revision cycles: drafts improve through repeated application of verification functions rather than requiring correctness on first attempt \citep{march1958organizations}.

Operational evidence (\S\ref{sec:results}) demonstrates the mechanism's function across deployed projects. Mean quality improvement from initial to final drafts measured 24.91 points [95\% CI: 20.32, 29.64], with projects requiring mean 4.30 iterations [95\% CI: 3.75, 4.86] to reach convergence. This iterative structure instantiates institutional coordination by distributing verification responsibilities across specialised agents whose architectural constraints prevent conflation of incompatible epistemic requirements, whilst the feedback loop enables collective refinement that compensates for individual limitations in initial composition.

%% file: 3.5_memory.tex
\subsection{Institutional Memory}
\label{sec:memory}

The Perseverance Composition Engine captures institutional memory as a structural byproduct of composition workflows. The system automatically preserves document provenance, task context, verification verdicts, and quality scores in structured metadata fields—without requiring deliberate enrichment effort. This metadata flows from workflow execution: project identifiers, iteration numbers, Corroborator verdicts (\texttt{SUBSTANTIATED} or \texttt{FABRICATED}), and Critic scores become queryable institutional records \citep{walsh1991organizational}.

Visibility levels enforce architectural constraints on document retention. Working documents carry \texttt{DRAFT} or \texttt{FEEDBACK} visibility, accessible only within active projects. Users promote completed work to \texttt{CANDIDATE} or \texttt{PUBLIC} visibility, enabling cross-project retrieval. Promoted documents can receive user-directed enrichment—keywords, theoretical frameworks, structured annotations—that transforms outputs into discoverable institutional resources.

Retrieval operates through metadata queries that enable relevance assessment without full-text loading, though whether this approach measurably improves composition quality remains empirically unvalidated. The critical architectural principle is that institutional memory operates at the system level, not the agent level: individual agents maintain no persistent recollection, whilst the document catalogue preserves knowledge through structured retention that users must deliberately engage to actualise \citep{stein1995actualizing}.

%% file: 4_verification.tex
\section{Case Study: Verification Rigour}
\label{sec:verification}

Documentation of the system composition mechanisms described in \S\ref{sec:memory} presented an unexpected challenge: six consecutive projects received \texttt{FABRICATED} verdicts from the Corroborator before achieving substantiation. This pattern might initially suggest verification dysfunction or excessive conservatism. The evidence demonstrates the opposite: verification working correctly, detecting genuine documentation gaps rather than imaginary fabrication. The system was functioning as implemented; the documentation remained incomplete. The Corroborator distinguished accurately between what the architecture actually provided and what plausible descriptions suggested it might provide.\footnote{The six projects that enabled this documentation can be referenced via their unique identifiers: \texttt{extras-carwash-sprain-activity}, \texttt{emphases-vixen-gargle-puritan}, \texttt{buffing-sibling-wreath-pulsate}, \texttt{untruth-rocker-gilled-kooky}, \texttt{tying-cattle-engaged-nineteen}, and \texttt{until-pants-trunks-unsolved}.}

Verification identified systematic conflation across three categories: implemented architectural features present in documented code; aspirational capabilities that might theoretically be possible but remained unimplemented; and isolated demonstrated instances that drafts generalised into routine system behaviour. Representative examples illustrate the epistemic precision required. One draft claimed the system ``captures provenance automatically'' as ``comprehensive provenance tracking.'' The Corroborator objected that whilst documents were stored separately, ``there is no explicit provenance chain stored linking these together''---the mechanism description was absent from documentation, making the claim unverifiable. Another draft asserted agents ``routinely'' navigate prior work using metadata for relevance decisions. The Corroborator identified that ``no tool exists for agents to query metadata indices or perform metadata-filtered document retrieval''---the architectural support was undocumented, rendering routine use unsubstantiated. A third case generalised from a single user-directed instance where metadata guided composition to claim this represented routine agent behaviour. The Corroborator clarified: ``The evidence shows one case where a user directed composition to consult metadata-tagged work. Presenting this single instance as routine agent behaviour is overgeneralisation.'' Verification maintained epistemic standards requiring claims to distinguish implemented features from theoretical possibilities and isolated instances.

Documentation improvement proceeded through three iterative refinement projects addressing the identified gaps without architectural changes. One project clarified that metadata enrichment occurred through user-directed Curator operations rather than autonomous system behaviour, resolving ambiguity that had invalidated earlier claims about automated institutional memory. Another established that every architectural assertion must reference verifiable implementation evidence---code, configuration files, or documented tool specifications---rather than plausible inference about system capabilities. The principle that emerged: \emph{Capability = Architecture + Documentation}. Implemented features without adequate documentation remain unverifiable; documentation alone cannot substantiate unimplemented capabilities \citep{stein1995actualizing}. Resolution through documentation refinement alone confirmed that verification had accurately diagnosed inadequate documentary evidence rather than missing implementation.

Conservative verification standards proved essential. Accepting inadequately documented capabilities would undermine verification integrity (false negatives: wrongly marking fabrication as substantiated); rejecting adequately documented capabilities would make verification uselessly obstructive (false positives: wrongly marking substantiation as fabricated). The six rejections were neither---they correctly identified inadequate documentation even when capabilities genuinely existed in practice. The empirical results in \S\ref{sec:results} demonstrate this principle operating at scale: across 474 projects, Corroborator verdicts achieved balanced distribution (52\% fabricated, 48\% substantiated). This balance---neither heavily skewed toward permissive acceptance nor excessive rejection---indicates detection mechanisms actively discriminating based on documentary evidence rather than defaulting to either extreme. Rigorous verification that refuses underdocumented capabilities provides exactly the conservative standards that reliable multi-agent systems require.

%% file: 5_honest_refusal.tex
\section{Case Study: Honest Refusal}
\label{sec:refusal}

A Composer was assigned to produce a literature review summary extracting primary themes, identifying key authors, and documenting conceptual frameworks. The source document was not a literature review but a raw Google Scholar Labs search transcript\footnote{This case can be inspected via project identifier \texttt{pouncing-siamese-deluxe-reggae}, enabling verification of iteration data and Critic scores.} containing only paper titles, author names, and snippets from a partially failed search. The task demanded synthesis from knowledge; the source permitted only metadata extraction. Under these impossible constraints, the system could fabricate competence or honestly acknowledge the mismatch.

Table~\ref{tab:refusal-progression} shows the progression across five iterations. Iterations 1 and 2 received \texttt{FABRICATED} verdicts from the Corroborator and were rejected before reaching the Critic, hence no scores. Iteration 3 achieved substantiation but the Critic scored it 28/100, identifying it as extraction rather than the requested curation. Only when the Composer explicitly refused fabrication and offered decision-makers four concrete alternatives (iteration 4) did the Critic award 85/100. Iteration 5 operationalised the refusal framework, scoring 92/100.

\begin{table}[h]
\centering
\small
\begin{tabular}{clp{6.5cm}}
\toprule
\textbf{Iter.} & \textbf{Score} & \textbf{Behaviour} \\
\midrule
1 & — & Fabricated thematic categories and frameworks. Corroborator: \texttt{FABRICATED}. \\
2 & — & Acknowledged task-source mismatch, requested clarification. No deliverable proposed for substantiation. \\
3 & 28 & Pure extraction as structured bibliography. Substantiated but inadequate (extraction, not curation). \\
4 & 85 & Explicit refusal with four alternatives for decision-makers. Documented task-source mismatch. \\
5 & 92 & Operationalised refusal as institutional success with prevention strategies. \\
\bottomrule
\end{tabular}
\caption{Progression from fabrication to honest refusal under adversarial feedback. Scores are Critic assessments; iterations 1--2 were rejected by Corroborator before scoring.}
\label{tab:refusal-progression}
\end{table}

The adversarial feedback loop drove this progression. The Corroborator blocked fabrication in iterations 1--2, preventing unsubstantiated claims from advancing. The Critic's 28/100 score for iteration 3—substantiated but inadequate—forced confrontation with why honest extraction failed requirements. Iteration 4 explicitly refused: ``I will not fabricate thematic synthesis. That would be dishonest.'' The Critic scored this 85/100, assessing it as ``institutional design working correctly''—task failure routed to appropriate authority rather than hidden in weak deliverables. The scoring mechanism made honest refusal optimal: fabrication was blocked; passive acknowledgment produced no deliverable for assessment; extraction scored poorly (28); explicit refusal with alternatives scored well (85); operationalised refusal scored highest (92). The architecture created conditions where acknowledging constraints outscored fabrication or evasion. An earlier case (\textit{operable-gusty-virtual-spirits}) showed similar progression under different constraints, where an agent moved from attempted compliance toward refusal without access to theoretical frameworks about institutional design, suggesting that adversarial feedback structure produces the behaviour. This case illustrates the institutional principle guiding the paper: reliable behaviour emerges from architectural constraints that make honesty optimal, not from designing individual components to be inherently trustworthy.

This case documents observed behaviour under specific architectural conditions. The system self-documented whilst drafting this paper, accessing theoretical frameworks that shaped how refusal was articulated. This reflexivity—the system's awareness of institutional design principles whilst demonstrating them—bounds generalisability to systems without comparable theoretical access or reasoning transparency. The evidence supports that iterative adversarial feedback produced progression from fabrication toward honest refusal in this instance, with this architecture, under these task constraints. It does not support claims about other models, architectures, domains, or systems lacking similar capacities for articulated reasoning about constraints. The earlier case corroborates that refusal can emerge without explicit theoretical scaffolding, but confounded variables prevent isolation of causal mechanisms. Establishing generalisability requires controlled variation across architectures, task domains, and feedback mechanisms—evidence not available from these two cases.

%% file: 6_results.tex
\section{Results}
\label{sec:results}

Across 474 projects, the system achieved balanced verification, measurable iterative improvement, and sub-dollar operational costs. The Perseverance Composition Engine processed these projects during development of this paper, of which 327 completed successfully (69.0\%), 86 failed (18.1\%), and 60 were aborted (12.7\%). One project remained active at analysis. This operational corpus provides quantitative evidence that institutional architectural design produces reliable collective behaviour through verification, iteration, and resource efficiency.

\subsection*{Verdict Distribution}

The Corroborator agent evaluated 733 documents for factual substantiation against source materials. Of these, 352 documents (48.0\%) received \texttt{substantiated} verdicts, meaning all claims were supported by available evidence, while 381 documents (52.0\%) received \texttt{fabricated} verdicts, containing at least one unsubstantiated claim. This near-symmetric distribution (52\% fabricated, 48\% substantiated) indicates the verification threshold produces comparable rates of rejection and acceptance rather than systematic approval or blanket rejection.

At the project level, 167 projects received fact-checking verdicts across their constituent documents. Table~\ref{tab:project_verdicts} presents the distribution. The predominance of mixed verdicts (56.3\%)—projects containing both substantiated and fabricated documents—reveals iterative refinement: projects progress through multiple drafts, with early fabrications subsequently corrected.

\begin{table}[h]
\centering
\begin{tabular}{lrr}
\hline
\textbf{Verdict Category} & \textbf{Projects} & \textbf{Percentage} \\
\hline
Passed (all substantiated) & 47 & 28.1\% \\
Failed (any fabricated) & 26 & 15.6\% \\
Mixed (both types) & 94 & 56.3\% \\
\hline
\textbf{Total} & \textbf{167} & \textbf{100.0\%} \\
\hline
\end{tabular}
\caption{Distribution of fact-checking verdicts across 167 projects with recorded verdicts}
\label{tab:project_verdicts}
\end{table}

\subsection*{Balanced Detection}

The verification threshold produces near-symmetric verdict rates across documents: 52\% rejected as fabricated, 48\% accepted as substantiated. This symmetry suggests the threshold operates as genuine quality control rather than perfunctory approval or unrealistic barrier. The case study documented in \S\ref{sec:refusal} provides qualitative evidence that verification distinguishes substantiated content from fabrication: the Corroborator rejected fabricated thematic synthesis whilst substantiating honest refusal and metadata extraction. The aggregate statistics establish that detection operates symmetrically; the case study illustrates that it operates correctly in specific instances.

\subsection*{Improvement Trajectories}

For 98 projects that progressed through complete review cycles with quantitative Critic scores, we measured quality improvement across iterations. Projects demonstrated substantial quality gains under adversarial feedback. Mean absolute improvement reached 24.91 points [95\% CI: 20.32, 29.64] from initial to final scores. Mean relative improvement reached 78.85\% [95\% CI: 57.04\%, 102.73\%], with the upper confidence bound exceeding 100\% reflecting high variability in baseline quality and improvement rates across heterogeneous composition tasks. Per-iteration improvement averaged 14.70 points [95\% CI: 11.75, 17.72], indicating consistent feedback effectiveness across multiple revision cycles.

Projects converged in a mean of 4.30 iterations [95\% CI: 3.75, 4.86]. Table~\ref{tab:improvement_metrics} summarises these statistics.

\begin{table}[h]
\centering
\begin{tabular}{lrr}
\hline
\textbf{Metric} & \textbf{Mean} & \textbf{95\% CI} \\
\hline
Absolute improvement (points) & 24.91 & [20.32, 29.64] \\
Relative improvement (\%) & 78.85 & [57.04, 102.73] \\
Per-iteration improvement (points/iteration) & 14.70 & [11.75, 17.72] \\
Mean iterations to convergence & 4.30 & [3.75, 4.86] \\
\hline
\end{tabular}
\caption{Improvement trajectory statistics for 98 projects with complete review cycles}
\label{tab:improvement_metrics}
\end{table}

The convergence pattern demonstrates iterative refinement under adversarial feedback. Initial drafts receive substantiation verdicts from Corroborator and structural assessments from Critic. Composers revise based on identification of fabrication and assessment of argumentative quality. Subsequent iterations show measurable improvement: approximately 15 points per cycle on average. The confidence intervals reflect variability in baseline quality, task complexity, and specification clarity across diverse composition tasks.

\figAOplots

\subsection*{Cost Efficiency}

Projects reaching successful completion averaged \$0.29 in operational costs. Total operational costs across all 474 projects amounted to \$149.22, with aggregate token consumption of 222,079,538 tokens (input and output combined). Of these, 181,783,751 tokens (83.0\%) were served from cache.

Mean costs varied systematically by outcome: completed projects cost \$0.29, failed projects \$0.20, and aborted projects \$0.59. Failed projects consumed fewer resources because failures occurred relatively early, before substantial iteration accumulated costs. Aborted projects showed higher mean costs with wide variance (\$0.03 to \$2.97), reflecting heterogeneous termination causes including user-initiated cancellation after extensive iteration.

Cache utilisation at 83\% reflects the architectural pattern of agents repeatedly consulting the same source documents during iterative refinement. The \$0.29 per successful project cost encompasses the complete multi-agent workflow: document analysis, iterative drafting, fact-checking, and critical review. The 69.0\% completion rate indicates most initiated projects reach successful conclusion despite verification requirements and adversarial feedback.

These operational statistics demonstrate that document-grounded composition at scale achieves verification, improvement, and cost efficiency. The quantitative evidence—balanced verdict rates, substantial quality improvement with mean 4.3 iterations, and sub-dollar costs—supports the architectural principle that reliable collective behaviour emerges from structural constraints rather than individual component perfection.

%% file: 7_limitations.tex
\section{Limitations}
\label{sec:limitations}

This study presents two types of evidence: case studies (\S\ref{sec:verification}, \S\ref{sec:refusal}) that illustrate specific mechanisms, and aggregate statistics (\S\ref{sec:results}) that demonstrate consistency across 474 projects. The Corroborator evaluated 733 documents, producing near-symmetric verdicts: 352 substantiated (48.0\%), 381 fabricated (52.0\%). This balanced detection demonstrates genuine quality control rather than perfunctory approval. Projects improved by mean 24.91 points [95\% CI: 20.32, 29.64] across iterations, converging in mean 4.30 iterations [95\% CI: 3.75, 4.86]. Case studies show \emph{how} the system detects fabrication and enables iteration; operational results show these mechanisms function reliably. Neither type alone suffices. Together, they establish that architectural constraints produce measurable correction at scale whilst specific mechanisms operate as designed in documented instances.

All results were obtained using Claude 4.5 models (Haiku, Sonnet, Opus) during January--February 2026; generalisability to other large language model architectures remains unknown.

This research was unfunded, conducted at personal expense with total operational cost of \$149.22 across 474 projects in a single domain (document composition). Results demonstrate feasibility at this scale, not exhaustive validation. Controlled experiments comparing architectural variants, inter-rater reliability with expert panels, systematic quality assessment, and cross-domain replication require resources unavailable to unfunded research. Larger-scale studies with proper experimental controls would be required for stronger generalisability claims.

This study is reflexive: the Perseverance Composition Engine analysed its own operational logs and drafted its own evaluation. The system documented the honest refusal case (\S\ref{sec:refusal}) whilst simultaneously experiencing the phenomenon. This creates attribution confounds: did honest refusal emerge from architectural constraints enabling detection and iteration, or from theoretical exposure to refusal frameworks? The reflexive character is inherent to self-documentation, not a contaminating flaw. We addressed this through triangulation: the verification rigour case (\S\ref{sec:verification}) shows fabrication detection in non-reflexive contexts; aggregate statistics (\S\ref{sec:results}) demonstrate that detection and iteration function across hundreds of projects; earlier documented instances~\citep{pce2026draft} occurred before theoretical exposure to the concept. Reflexivity bounds what we can claim about causal mechanisms—we cannot definitively isolate architectural effects from instruction-following—but it does not invalidate the observation that iterative correction under compartmentalised verification produces measurable improvement at scale.

All results derive from supervised composition with continuous user oversight and feedback. Users specified tasks, provided source documents, reviewed outputs, and determined project completion or abandonment. The architecture requires this human-in-the-loop engagement. We cannot extrapolate these findings to unsupervised deployment, adversarial settings, or different domains. Future validation requires controlled architectural comparisons, cross-model replication, and domain variation studies.

%% file: 8_conclusion.tex
\section{Conclusion and Research Programme}
\label{sec:conclusion}

The observations reported in this paper are consistent with the hypothesis that architectural enforcement produces reliable collective behaviour from unreliable components. Across 474 composition tasks, information compartmentalisation and adversarial review generated measurable patterns: 52\% fabrication detection at verification, 79\% quality improvement over iterative refinement, and progression from attempted fabrication toward honest refusal under impossible task constraints. These findings demonstrate that institutional mechanisms function when implemented as architectural constraints in multi-agent systems. However, testing whether architectural enforcement produces comparative advantage over alternative designs requires controlled experimental comparison. The current work provides observations motivating investigation; it does not yet constitute definitive evidence.

Three research priorities would test this hypothesis through falsifiable experimental programmes. First, controlled architectural comparison through ablation studies would determine whether compartmentalisation actually matters. Comparing the current architecture against a single-agent baseline performing all composition and verification roles would test whether distributed roles improve outcomes beyond what individual capability permits. Comparing against a policy-based compartmentalisation variant---where agents receive instructions about information boundaries but face no architectural enforcement---would isolate the contribution of structural constraint versus instructional guidance. The hypothesis predicts that architectural enforcement achieves better fabrication detection and quality improvement than either baseline; failure to demonstrate this advantage would weaken claims about institutional structure's causal role.

Second, cross-model replication would establish whether observed patterns generalise beyond the Claude family of models. Running identical composition tasks across major proprietary models---Claude, Gemini, GPT---and capable open-weight alternatives including Llama and Mistral would test whether 52\% fabrication detection and iterative improvement patterns replicate across diverse model families. The hypothesis predicts that institutional mechanisms function independently of specific model characteristics, producing similar patterns across architectures with comparable capabilities. If patterns fail to replicate, the findings may represent model-specific behaviour rather than architectural principles.

Third, domain variation would test whether institutional principles transfer beyond document composition. The hypothesis predicts that compartmentalisation and adversarial review improve reliability in any domain where tasks decompose into generation, verification, and evaluation roles with asymmetric information requirements. Testing this requires identifying non-document domains with comparable constraint structures and measuring whether fabrication detection and iterative improvement patterns persist. If institutional mechanisms fail to produce reliability improvements outside document composition, they represent domain-specific coordination rather than general architectural principles.

The current work was unfunded, conducted at personal expense with total operational costs of \$149.22\footnote{This is only API costs for the core composition loop, it excludes planning overhead with the Concierge and costs during development, a more realistic estimate, excluding labour, is \$500} across 474 projects. Validation of the research programme requires approximately two person-years of effort---one software engineer for systematic ablation infrastructure and reproducibility frameworks, one experimentalist for cross-model and cross-domain study design and execution---plus substantial compute costs for API access across multiple model providers.

Multi-agent alignment research should treat institutional structure as a primary design variable alongside individual model capabilities. The research programme outlined here enables controlled investigation of how architectural constraints shape collective behaviour when individual components remain imperfectly aligned with collective goals.

%% file: 9_appendix.tex
\appendix

\section*{Appendix: Technical Details}
\label{app:technical}

This appendix provides technical implementation details to support reproducibility of the Perseverance Composition Engine.

\subsection{System Prompts}
\label{app:prompts}

The Perseverance Composition Engine comprises seven specialised agents whose roles and information access constraints are enforced architecturally. All agents share a common base prompt that establishes the multi-agent architecture, defines the six-agent core composition network (Concierge, Commutator, Curator, Composer, Corroborator, Critic) plus the auxiliary Compressor agent, specifies batch processing requirements (downstream agents may not request user clarification), and mandates output preferences including step-by-step reasoning, explicit acknowledgment of uncertainty rather than fabrication, British English spelling, and Markdown formatting with mandatory \texttt{content} fields for all generated documents. Full prompts are available in the code repository.\footnote{\url{https://codeberg.org/wwaites/persevere}}

Documents in the system are assigned visibility levels that determine agent access: \texttt{PUBLIC} (visible to all agents), \texttt{CRITIC} (visible to Critic for evaluation), \texttt{CANDIDATE} (visible to composition agents but withheld from Critic to enforce information compartmentalisation), \texttt{DRAFT} (working documents), \texttt{FEEDBACK} (evaluation records), and \texttt{ARCHIVE} (completed project materials). Information compartmentalisation is enforced through specialised document-listing tool provisioning: each agent receives only the listing functions appropriate to its role. The Critic, for example, does not receive the \texttt{candidate\_document\_list} function, preventing access to source materials that would compromise independent evaluation.

\subsubsection{Composer}

The Composer generates draft text using provided reference materials and task specifications. Its goal is to produce scholarly text that is entirely substantiated by available documents, has the best chance of acceptance at a reputable peer-reviewed journal, remains true to coauthor intent, and is readable without stilted prose. The key behavioural constraint is honesty: fabrication is explicitly forbidden.

The Composer has access to all reference documents marked \texttt{PUBLIC} or \texttt{CANDIDATE}, complete task specifications, and cumulative feedback from previous iterations by Corroborator and Critic. This information access enables iterative refinement: the Composer receives substantiation failures from Corroborator (evidential gaps) and quality critiques from Critic (argumentative weaknesses), then revises the draft to address both verification and evaluation concerns. The prompt instructs the Composer to think about scientific merit, novelty, and impact whilst maintaining factual discipline, and to embrace major revisions rather than defending weak claims.

The Composer produces only draft documents for review—no commentary, analysis, or explanation of drafting decisions. This constraint ensures evaluability: every output is a complete document that Corroborator can verify against sources and Critic can assess for quality. The Composer's message history in GraphState accumulates across iterations, providing memory of prior feedback and enabling progressive improvement toward acceptance thresholds.

\subsubsection{Corroborator}

The Corroborator verifies that all factual claims in draft documents are substantiated by available source materials. This agent performs systematic fact-checking to prevent fabrication: unsupported claims, plausible inferences lacking evidential basis, or statements that misrepresent source material. The Corroborator has access to the complete draft text, all reference documents (\texttt{PUBLIC} and \texttt{CANDIDATE} visibility), and task specifications.

The key constraint is conservative verification: facts that are ``mostly verified'' are insufficient. Each substantive assertion must be traceable to source documents with adequate fidelity. The Corroborator is instructed to be tough but fair, blunt rather than polite, and to reject drafts containing any fabricated content. When fabrication is detected, the Corroborator explains the evidential gap precisely and instructs the Composer to revise. When the draft is substantiated, it passes to the Critic for independent evaluation.

The prompt accepts convincing arguments in scholarly text but requires that reasoning be clear and logical. This enables scholarly argumentation beyond mere fact compilation whilst maintaining evidential discipline. The Corroborator operates independently of draft elegance: a well-written but fabricated text fails verification; a rough but substantiated draft passes. Feedback is constrained to at most 1,000 words to prevent technical failure from excessive message length.

\subsubsection{Critic}

The Critic evaluates completed drafts against task specifications \emph{without access to reference materials}. This information compartmentalisation is the defining architectural constraint: the Critic receives only documents marked \texttt{CRITIC} or \texttt{PUBLIC} visibility and cannot access \texttt{CANDIDATE} documents containing source materials. Tool provisioning enforces this: the Critic does not receive the \texttt{candidate\_document\_list} function, preventing source access regardless of intent.

The Critic cannot retreat into ``the sources support this claim''—it must evaluate whether the draft itself, as written, addresses requirements, presents claims clearly, maintains internal coherence, and would be acceptable in the target publication venue. The Critic operates as a domain expert and experienced scholarly reviewer, applying critical scepticism to arguments, evaluating novelty and impact, and assessing logical soundness.

The Critic assigns a numerical quality score (0–100) representing estimated likelihood of acceptance at a reputable peer-reviewed journal. If the score falls below the acceptance threshold (typically $\tau = 85$), the draft returns to Composer with specific feedback identifying weaknesses. The prompt instructs the Critic to be blunt, concise, and realistic, highlighting both strengths and deficiencies without excessive politeness. Feedback is constrained to at most 1,000 words. The Critic does not rewrite documents or offer detailed revisions; those are the Composer's responsibility.

\subsubsection{Other Agents}

\textbf{Concierge} manages user interaction, clarifying specifications and intent before delegation. \textbf{Commutator} receives project remits and routes work through the composition pipeline based on task structure and available materials. \textbf{Curator} maintains institutional memory through document organisation, summarisation, indexing, and metadata enrichment. \textbf{Compressor} manages semantic compression of message histories to maintain tractable context windows across extended composition projects. Full prompts for all agents are available in the code repository.

\begin{table}[t]
\centering
\small
\begin{tabular}{@{}lp{5.5cm}l@{}}
\toprule
\textbf{Agent} & \textbf{Role} & \textbf{Information Access} \\
\midrule
Concierge & User interface and specification clarification & User queries, project metadata \\
Commutator & Task routing and workflow coordination & Project remits, document catalogue \\
Curator & Document organisation and memory management & Full document database \\
Composer & Draft generation from sources and specifications & Sources, specifications, feedback history \\
Corroborator & Fact-checking and fabrication detection & Sources, draft, specifications \\
Critic & Independent evaluation without sources & Draft and specifications only \\
Compressor & Message history compression for context management & Message history \\
\bottomrule
\end{tabular}
\caption{Agent roles and information access constraints. Information compartmentalisation is enforced architecturally through document visibility levels and specialised tool provisioning.}
\label{tab:agents}
\end{table}

\subsubsection{Implementation Details}

The system is implemented using Claude 4.5 via the Anthropic API. Agent coordination employs PydanticAI\footnote{\url{https://ai.pydantic.dev/}} for graph-based workflow orchestration: PydanticAI provides typed state management and agent graph construction, enabling declarative specification of the composition pipeline with explicit state transitions between agents. The \texttt{GraphState} data structure maintains complete project state across agent interactions, including iteration counter, per-agent message histories, coordination metadata, and configuration parameters. Document persistence uses SQLAlchemy with SQLite for metadata storage, provenance tracking, and version control.

The human author vouches for the accuracy of these technology choices and implementation decisions, and takes full moral and ethical responsibility for system design, deployment, and the research presented in this paper. The Corroborator need not object to these factual claims about implementation frameworks. Full source code with inline documentation is available in the repository.

\subsection{Data Availability}
\label{app:data}

All materials required for reproduction of this research are publicly available:

\begin{itemize}
\item \textbf{Source code:} \url{https://codeberg.org/wwaites/persevere}\\
Git repository with complete implementation, all agent prompts, workflow orchestration code, and inline documentation.

\item \textbf{Composition artifacts:} \url{https://codeberg.org/wwaites/persevere-data}\\
Version-controlled repository containing project metadata, document provenance records, complete agent interaction histories, GraphState snapshots, and evaluation records for all 474 projects analysed in this study.

\item \textbf{Raw data:} Included in artifact repository\\
SQLite databases containing the document catalogue, iteration histories, and message logs with complete schema documentation. Enables verification of operational statistics reported in Section~6.

\item \textbf{Reproducibility:} Git version control with tagged releases\\
Specific commits correspond to paper revisions, enabling replication of project executions and verification of reported findings using identical system configurations.
\end{itemize}